\documentclass[10pt,twocolumn,letterpaper]{article}

\usepackage{wacv}
\usepackage{times}
\usepackage{epsfig}
\usepackage{graphicx}
\usepackage{amsmath}
\usepackage{amssymb}

\usepackage{booktabs}
\usepackage{floatrow}
\usepackage{subcaption} 
\usepackage{textcomp}
\usepackage{stackengine}
\usepackage{resizegather}

\usepackage{color}

\newcommand\blfootnote[1]{%
  \begingroup
  \renewcommand\thefootnote{}\footnote{#1}%
  \addtocounter{footnote}{-1}%
  \endgroup
}
\cvprfinalcopy
\def\cvprPaperID{6043} % *** Enter the wacv Paper ID here

\ifcvprfinal\fi
\begin{document}

%%%%%%%%% TITLE
\title{From Third Person to First Person:\\Dataset and Baselines for Synthesis and Retrieval}

% Authors at the same institution
%\author{First Author \hspace{2cm} Second Author \\
% Authors at different institutions

\author{
Mohamed Elfeki$^*$,Krishna Regmi$^*$, Shervin Ardeshir, and  Ali Borji\\
% \vspace*{-0.35cm}
University of Central Florida, Center for Research in Computer Vision (CRCV)\\
% $^{2}$Netflix Research, LA CA\\
{\tt\small \{elfeki, regmi, ardeshir\}@cs.ucf.edu, aliborji@gmail.com}
\vspace*{-0.5cm}
}

\maketitle

\begin{abstract}
First-person (egocentric) and third person (exocentric) videos are drastically different in nature. The relationship between these two views have been studied in the recent years, however, it has yet to be fully explored. In this work, we introduce two datasets (synthetic and natural/real) containing simultaneously recorded egocentric and exocentric videos. We also explore relating the two domains (egocentric and exocentric) in two aspects. First, we synthesize images in the egocentric domain from the exocentric domain using a conditional generative adversarial network (cGAN). We show that with enough training data, our network is capable of hallucinating how the world would look like from an egocentric perspective, given an exocentric video. Second, we address the cross-view retrieval problem across the two views. Given an egocentric query frame (or its momentary optical flow), we retrieve its corresponding exocentric frame (or optical flow) from a gallery set. We show that using synthetic data could be beneficial in retrieving real data . We show that performing domain adaptation from the synthetic domain to the natural/real domain, is helpful in tasks such as retrieval. We believe that the presented datasets and the proposed baselines offer new opportunities for further research in this direction. The code and dataset are publicly available.\blfootnote{* equal contributions}\footnote{www.github.com/M-Elfeki/ThirdToFirst}
\end{abstract}

\section{Introduction}
\label{sec:introduction}
Recently egocentric cameras have gathered a plethora of data and have provided the opportunity to study first person vision extensively. At the same time, tremendous amount of research has been conducted on more traditional types of videos collected using static third-person cameras. We refer to these videos as exocentric. 
First-person and third-person domains, although drastically different, can be related together. In this work we take a step towards exploring this relationship. We are motivated by the fact that research in exocentric domain has a longer history relative to the first-person domain. Hence, there are more available datasets and benchmarks in this domain. Thus, effective transfer of information from third person to first person perspective could be very beneficial to research in the first-person domain. Understanding the relationship between these domains will facilitate exploiting existing models and solutions in exocentric domain and applying them to similar problems in egocentric domain. 

\begin{figure}[t]
\begin{center}
\includegraphics[width=\linewidth]{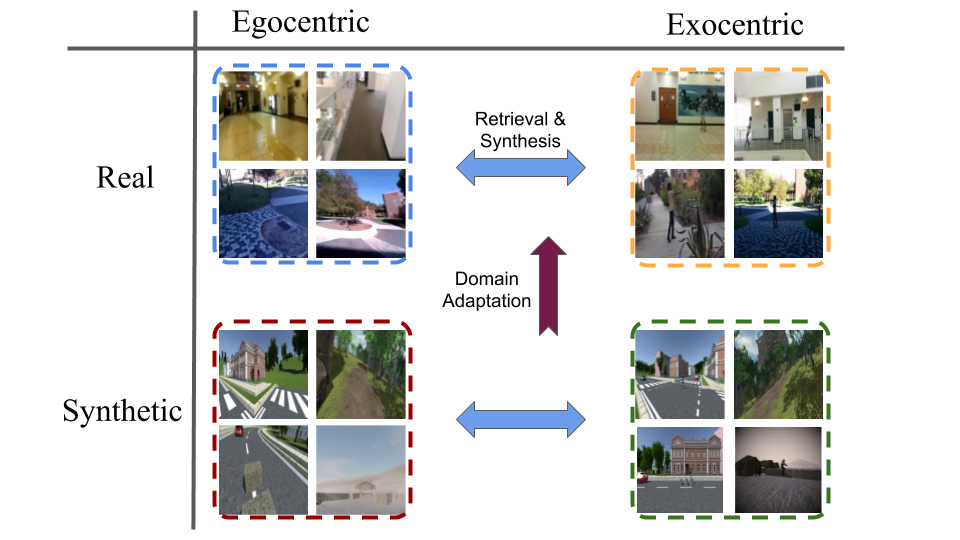}
\caption{We explore the transformation between two viewpoints: egocentric and exocenteric. Learning this transformation tends to be non-trivial, therefore we adapt the learning from synthetic to real data. We study this transformation by performing two cross-domain tasks: retrieval and synthesis.}%We show few frames from egocentric and exocentric domains for real and synthetic datasets. We exploit to study the relationship between these two domains. Domain adaptation from synthetic to real is helpful when we have less real dataset which is difficult to collect compared to synthetic dataset.}
\label{fig:teaser}
\end{center}
\end{figure}

Our contributions in this work are three folds as explained below.\\
\noindent\textbf{Dataset:} We collect two datasets (synthetic and real), each containing simultaneously recorded egocentric and exocentric video pairs, where the egocentric is captured by body mounted cameras and the exocentric is captured by static cameras, capturing the egocentric camera holders performing diverse actions covering a broad spectrum of motions. We collect a large scale synthetic dataset generated using game engines, and provide frame level annotation on egocentric and exocentric camera poses, and the actions being performed by the actor. We also collect a smaller scale dataset of simultaneously recorded real egocentric and exocentric videos of actors performing different actions. We believe that the datasets and the annotations will be useful for exploring the relationship between first and third person videos in many aspects such as video retrieval and synthesis (as we explore here), action recognition, pose estimation, and 3D reconstruction. We believe that simultaneously recorded egocentric and exocentric videos could be beneficial in effectively exploring the relationship between these two domains, and could be benificial to the community. \\ 
\noindent\textbf{Image Synthesis:} Given an exocentric side-view image, we aim to generate an egocentric image hallucinating how the world would look like from a first person perspective. Synthesis is a very challenging computer vision problem, especially when the generation is conditioned on images with drastically different views. In our work, the images in two domains often do not have a significant overlap in terms of their fields of view. Thus, transforming the appearances across the two views is non-trivial. As one of the contributions of this work, we attempt to address this problem across third person and first person images using conditional generative adversarial networks.\\
\noindent\textbf{Retrieval:} Given an exocentric frame in a video or its momentary optical flow (with respect to the previous frame), we explore retrieving its corresponding egocentric frame (or optical flow). To do so, we train a two stream convolutional neural network seeking a view invariant representation across the two views given a momentary optical flow map (a 2 channel input). We also train another network for RGB values (a 3 channel input). We perform domain adaptation across synthetic and real domain and show that using synthetic data improves the retrieval performance on real data.

In the following, we cover the related works in section \ref{sec:relatedWork}, describe the datasets in section \ref{sec:dataset}, our framework in section \ref{sec:framework}, and experimental results in section \ref{sec:experimentalResults}. Finally, we conclude the work in section \ref{sec:conclusion}.

% \vspace{-10pt}
\section{Related Work}
\vspace{-5pt}
\label{sec:relatedWork}
\noindent\textbf{Egocentric Vision:}
First person vision, a.k.a egocentric vision, has become increasingly popular in the computer vision community. A lot of research has been conducted in the past few years ~\cite{egoKanade,egoEvolutionSurvey}, including object detection \cite{egoObjectDetection}, activity recognition \cite{egoDailyAction,egoActionFathi} and video summarization \cite{egoVideoSummarization}. Motion in egocentric vision, in particular, has been studied as one of the fundamental features of first person video analysis. Costante et al. \cite{costante2016exploring} explore the use of convolutional neural networks (CNNs) to learn the best visual features and predict the camera motion in egocentric videos. Su and Grauman \cite{su2016detecting} propose a learning-based approach to detect user engagement by using long-term egomotion cues. Jayaraman et al. \cite{Jayaraman_2015_ICCV} learn the feature mapping from pixels in a video frame to a space that is equivariant to various motion classes. Ma et al. \cite{ma2016going} have proposed a twin stream network architecture to analyze the appearance information and the motion information from egocentric videos and have used these features to recognize egocentric activities. Action and activity recognition in egocentric videos have been hot topics in the community. Ogaki et al. \cite{6239188} jointly used eye motion and ego motion to compute a sequence of global optical flow from egocentric videos. Poleg et al. \cite{poleg2016compact} proposed a compact 3D Convolutional Neural Network (3DCNN) architecture for long-term activity recognition in egocentric videos and extended it to egocentric video segmentation. Singh et al. \cite{Singh_2016_CVPR} used CNNs for end-to-end learning and classification of actions by using hand pose, head motion and saliency map. Li et al. \cite{Li_2015_CVPR} used gaze information, in addition to these features, to perform action recognition.
In their work, Matsuo et al. \cite{Matsuo_2014_CVPR_Workshops} have proposed an attention based approach for activity recognition by detecting visually salient objects.\\

\noindent\textbf{Relating first and third person videos:} The relationship between egocentric and top-view information has been explored in tasks such as human identification \cite{ardeshir2016ego2top,ardeshir2018integrating,fan2017identifying}, semantic segmentation\cite{ardeshir2015geo} and temporal correspondence\cite{ardeshirEgocentricMeets}. In this work, we relate two different views of a motion, which can be considered as a knowledge transfer or domain adaptation task. Knowledge transfer has been used for the multi-view action recognition (e.g., \cite{junejo2008cross,liu2011cross,li2012discriminative}) in which multiple exocentric views of an action are related to each other. Having multiple exocentric views allows geometrical and visual reasoning, since: a) the nature of the data is the same in different views, and b) the actor is visible in all cameras. In contrast, our paper aims to automatically learn mappings between two drastically different views, egocentric and exocentric. To the best of our knowledge, this is the first attempt in relating these two domains for transferring motion information.
Cross-view relations have also been studied between egocentric (first person) and exocentric (surveillance or third-person) domains for action classification. \cite{DBLP:conf/accv/SoranFS14} utilize the information from one egocentric camera and multiple exocentric cameras to solve the action recognition task, and \cite{ardeshir2018exocentric} learns a mapping between first person and third person actions. \\

\noindent\textbf{Generative Adversarial Networks:} Goodfellow et al. \cite{goodfellow2014generative} proposed the initial version of Generative Adversarial Networks for generating realistic images. Prior to that, Restricted Boltzmann Machines \cite{Hinton:2006:FLA:1161603.1161605,Smolensky:1986:IPD:104279.104290} and deep Boltzmann Machines \cite{salakhutdinov2009deep} have been used for that purpose. GANs have been used in conditional settings to synthesize images controlled by different parameters, such as labels of digits \cite{DBLP:journals/corr/MirzaO14}, images~\cite{pix2pix2017, regmi2018cross, Regmi2018CrossviewIS}, textual descriptions~\cite{pmlr-v48-reed16,han2017stackgan}. GANs are exploited for inpainting tasks by ~\cite{pathak2016context,yeh2017semantic}. We are the first to synthesize cross-view images involving egocentric and exocentric domains. In this work, we condition the generative adversarial networks on exocentric view image and attempt to hallucinate how the world looks from egocentric perspective.

\section{Dataset}
\label{sec:dataset}
We collect a real dataset and a synthetic dataset containing simultaneously recorded egocentric and exocentric videos. In what follows, we briefly describe the two datasets and their statistics.
% \vspace{-10pt}
\subsection{Real Dataset}
\begin{figure}[t]
\begin{center}
\includegraphics[width=1\linewidth]{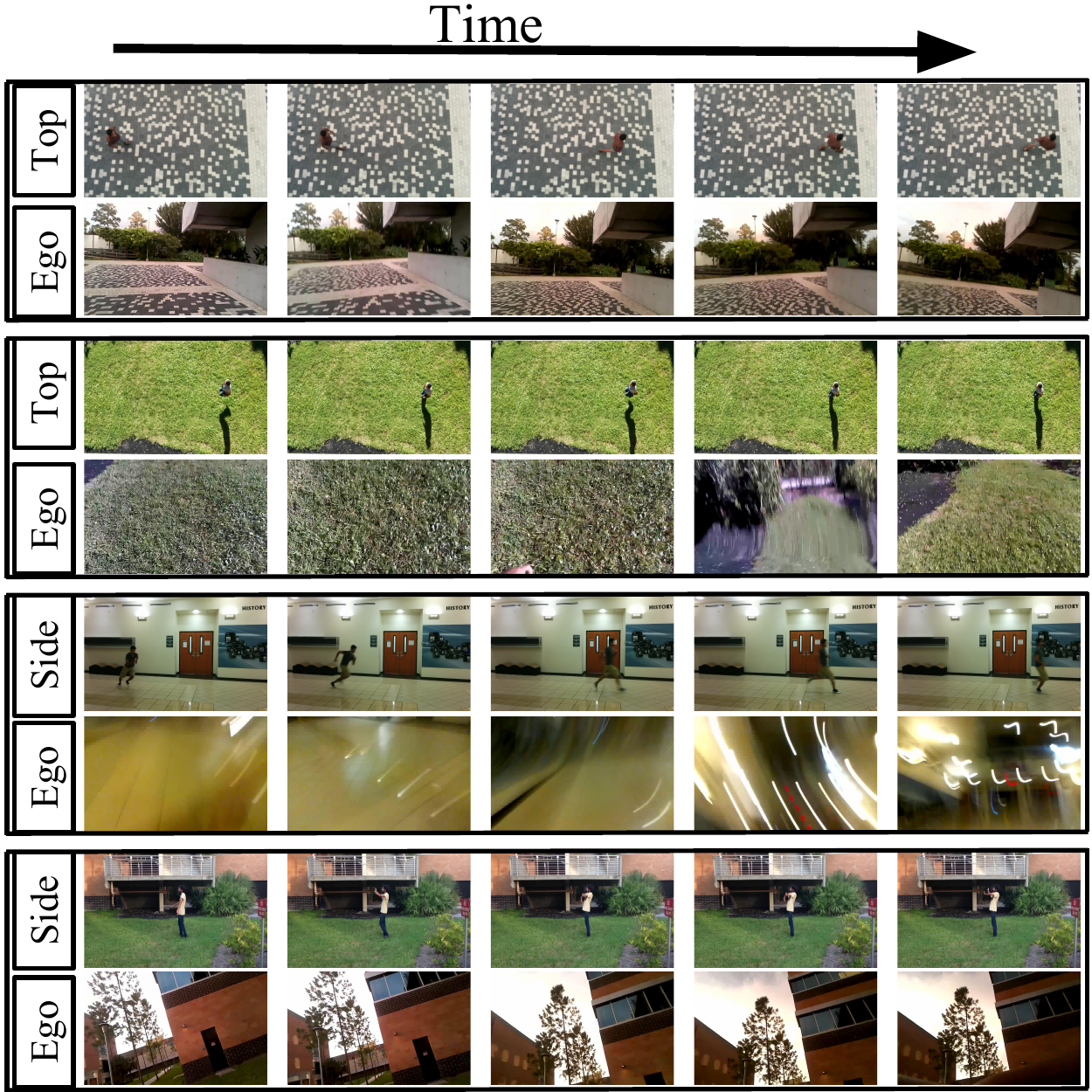}
% \vspace{-15pt}
\caption{Examples from the real dataset: simultaneously recorded Ego-Top and Ego-Side pairs are shown.}
\label{fig:realDataExamples}
\end{center}
\end{figure}

% \vspace{-5pt}
We present a dataset containing simultaneously recorded egocentric and exocentric videos covering a wide range of first and third person movements and actions. As this dataset is designed for studying the relationship between these two views, we isolate the egocentric camera holder in the third person video and thus, collect videos in which there is only a single person collecting an egocentric video and being recorded by an exocentric video. We collect a dataset containing 531 video pairs. Each video pair contains one egocentric and one exocentric (side or top-view) video. The pair of videos are temporally aligned, which will provide corresponding ego-exo image pairs. Some example frames are shown in Fig. \ref{fig:realDataExamples}. Each pair is collected by asking an actor to perform a range of actions (walking, jogging, running, hand waving, hand clapping, boxing, and push ups) covering a broad range of various motions and poses in front of an exocentric camera (top or side view), while wearing an egocentric body-worn camera capturing the actor's motion from the first person perspective. Details about the number of videos and statistics for training and testing are included in Table \ref{tab:realDataset}. 
% \vspace{-15pt}
\subsubsection{Metadata and Annotations.} We provide frame level action labels for the videos in each view. Actions consist of walking, jogging, running, waving, boxing, clapping, jumping, and doing push-ups.

\begin{table*}[h!]
\begin{center}
  \begin{tabular}{ l | c | c | c | c | c | c | c | c | }
    %\hline
 & \multicolumn{2}{c|}{Training Pairs} & \multicolumn{2}{c|}{Validation Pairs} & \multicolumn{2}{c|}{Testing Pairs} & \multicolumn{2}{c|}{Total Number of Pairs}  \\ 
  \cline{2-9}
 %\hline
     & $\#$Vid & $\#$Frames & $\#$Vid & $\#$Frames & $\#$Vid & $\#$Frames  & $\#$Vid & $\#$Frames \\ \hline    
    Ego-Side & 124 & 26,764 & 61 & 13,412 & 70 & 13,788& 255 & 53,964 \\ \hline
    Ego-Top & 135 & 28,408 & 68 & 12,904 & 73 & 14,064 & 276 & 55,376 \\
    \hline
  \end{tabular}  
\end{center}
% \vspace{-12pt}
\caption{Details of Real Dataset in terms of the number of training, validation and testing video and frame pairs.}
\label{tab:realDataset}
\vspace{-5pt}
\end{table*}

% \vspace{-35pt}

\begin{table*}[h!]
\begin{center}
  \begin{tabular}{ l | c | c | c | c | c | c | c | c | }
 & \multicolumn{2}{c|}{Training Pairs} & \multicolumn{2}{c|}{Validation Pairs} & \multicolumn{2}{c|}{Testing Pairs} & \multicolumn{2}{c|}{Total Number of Pairs}  \\ 
  \cline{2-9}
     & $\#$Vid & $\#$Frames & $\#$Vid & $\#$Frames & $\#$Vid & $\#$Frames  & $\#$Vid & $\#$Frames \\ \hline    
    Ego-Side & 208 & 119,115 & 109 & 6,702 & 95 & 6,778& 412 & 132,595 \\ \hline
    Ego-Top & 208 & 119,115 & 109 & 6,702 & 95 & 6,778& 412 & 132,595 \\
    \hline
  \end{tabular}  
\end{center}
% \vspace{-15pt}
\caption{Details of Synthetic Dataset in terms of the number of training, validation and testing video and frame pairs.}
\label{tab:syntheticDataset}
\end{table*}

% \vspace{-20pt}
\subsection{Synthetic Data}
\begin{figure}[t]
\begin{center}
\includegraphics[width=1\linewidth]{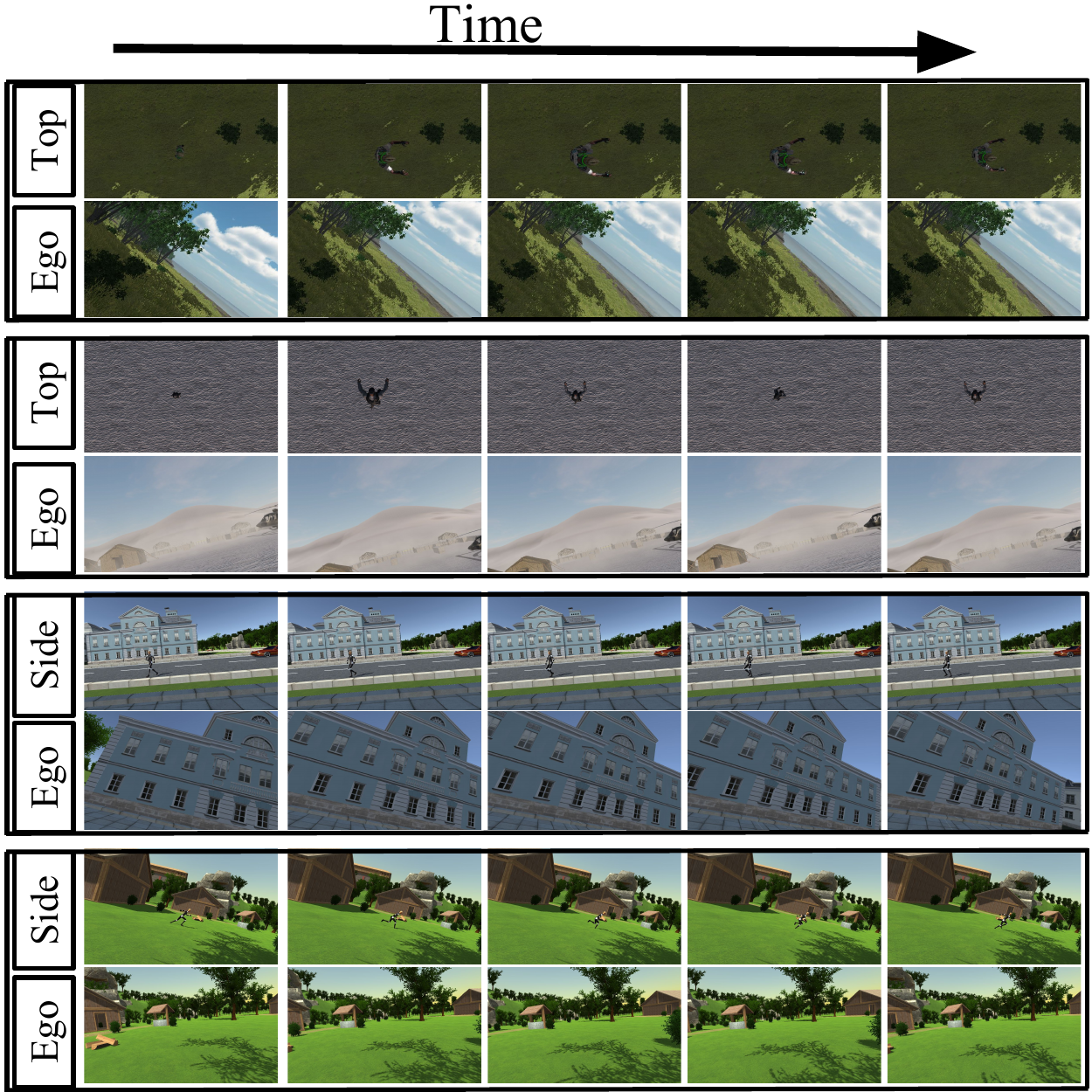}
% \vspace{-15pt}
\caption{Examples from the synthetic dataset: simultaneously recorded Ego-Top and Ego-Side pairs are shown.}
\label{fig:syntheticDataExamples}
\end{center}
\end{figure}

Since simultaneously recorded egocentric and exocentric videos are not abundant, collecting such data from the web and in large scale is not feasible. In order to attain a large number of samples, we collect a synthetic dataset using graphics engines. Several environments and actors were used in unity 3D platform, programmed to perform actions such as walking, running, jumping, crouching, etc. A virtual egocentric camera was mounted on the actor's body, while static virtual top/side view cameras were also positioned in the scene. We collected a large number of examples (more than 130,000 frames per camera) of such data. A few examples are shown in Fig. \ref{fig:syntheticDataExamples}. In order to add variation to the data and make it resemble real data, we added slight random rotations to the virtual cameras.
% \vspace{-15pt}

In our synthetic dataset, we have a total of 4 environments with 5, 7, 10 and 10 scenes. Scene refer to a location where the actions are recorded. For each environment, we use two scenes for testing and the rest for validation and training.

\subsubsection{Metadata and Annotations.} We provide frame level action labels, along with egocentric and exocentric camera poses. The action classes consist of walking, running, crouching, strafing, and jumping. 

% \vspace{-5pt}
\noindent\textbf{Dataset Value:} We believe that the relationship across views (egocentric and exocentric) and modalities (synthetic and real data) could be explored in many aspects. Given that the dataset contains simultaneously recorded videos, and it contains frame level annotations in terms of action labels and camera poses, we believe that it could be used for many tasks such as video retrieval and video synthesis, for which we provide some baselines. Also this relationship could be explored in other tasks such as action recognition, camera pose estimation, human pose estimation, 3D reconstruction, etc.

% \vspace{-10pt}
\section{Framework}
\label{sec:framework}
\subsection{Image Synthesis}
Generative Adversarial Networks~\cite{goodfellow2014generative} are useful in synthesizing natural looking images which are not possible by minimizing the pixel-wise loss only during the training. GANs employ a generator network (G) that synthesizes the images very close to the training data distribution from noise distribution and a discriminator network (D) that is trained to discriminate between the samples generated by $G$ and the original samples from the true data distribution. The discriminator acts as a learnable loss function to the generator to improve realism in synthesized images. 

% The network architecture consists of two components: a generator and a discriminator, which are trained jointly. Generator $G$ maps a noise vector to a sample which fits the training data distribution. Discriminator $D$ is trained to discriminate between the samples generated by $G$ and the original samples from the true data distribution. The objective function is formalized as: 

% \begin{equation}
% \stackanchor{min }{G}  \stackanchor{max  }{D} L_{GAN} (G,D)  =  E_{x\sim p_{data}(x)} [log D(x) ]+ E_{z\sim p_z(z)}[log(1 - D(G(z)))],
% \end{equation}
% $x$ is real data sampled from data distribution ${p_{data}}$ and $z$ is a $d$-dimensional noise vector sampled from a Gaussian distribution ${p_{z}}$. 

Conditional GANs use an auxiliary variable (e.g., labels~\cite{DBLP:journals/corr/MirzaO14}, text embeddings~\cite{pmlr-v48-reed16,han2017stackgan} or images~\cite{pix2pix2017, regmi2018cross, Regmi2018CrossviewIS, CycleGAN2017,pmlr-v70-kim17a}) as input to synthesize samples. Both $G$ and $D$ are shown the conditioning variable. $G$ generates the target image using the auxiliary input. The conditioning variable is paired with real/synthesized image and shown to $D$ and $D$ makes its prediction of whether the image pair it sees is real or fake.

Earlier works in GAN~\cite{pix2pix2017,regmi2018cross,pathak2016context} used $L1$ or $L2$ distances between real and generated image pairs as additional term in loss function to encourage the generator to synthesize samples similar to the ground truth. Here, we use $L1$ distance as it increases image sharpness in the generation tasks. 

% For conditional GAN, the objective function is defined as:
% \vspace{-5pt}

% \begin{equation}\label{eq_cond}
% L_{cGAN}(G,D) = E_{x,c\sim p_{data}(x, c)} [log D(x,c)] + E_{x', c \sim p_{data}(x',c)}[ log(1 - D(x',c))],
% \end{equation}
% where $x'$ = $G(z,c)$ is the synthesized image.

% % \vspace{-5pt}

% \begin{equation}\label{eq_cond_l1}
% \stackanchor{min  }{G} L_{L1}(G)=E_{x,x'\sim p_{data}(x,x')}[\mid \mid x - x' \mid \mid _1],
% \end{equation}

% The objective function for such conditional GAN network is the sum of equations \eqref{eq_cond} and~\eqref{eq_cond_l1}.

In this work, we use an exocentric image $(I_{exo})$ as a conditional input to synthesize the ego image $(I_{ego})$. We minimize the adversarial loss and L1 loss during training. 
The conditional GAN loss and $L1$ loss are represented by Eq. \eqref{eq_gan} and Eq. \eqref{eq_l1}, respectively.

% \begin{equation}\label{eq_gan}
% L_{cGAN}(G,D) = E_{I_{ego},I_{exo} \sim p_{data}(I_{ego},I_{exo})} [log D(I_{ego},I_{exo})] + E_{I_{exo}, I_{ego}' \sim p_{data}(I_{exo}, I_{ego}')}[ log(1 - D(I_{ego}',I_{exo}))],
% \end{equation} 

\begin{equation}\label{eq_gan}
\begin{split}
\resizebox{0.8\hsize}{!}{$%
\stackanchor{min }{G} \stackanchor{max  }{D} L_{cGAN}(G,D) = E_{I_{ego},I_{exo} \sim p_{data}(I_{ego},I_{exo})} [log D(I_{ego},I_{exo})]
$%
}%
\\ \hspace*{0.5in}
\resizebox{0.6\hsize}{!}{$%
 + E_{I_{exo}, I_{ego}' \sim p_{data}(I_{exo}, I_{ego}')}[ log(1 - D(I_{ego}',I_{exo}))],
$%
}%
\end{split}
\end{equation}

\begin{equation}\label{eq_l1}
\stackanchor{min  }{G} L_{L1}(G)=E_{I_{ego},I_{ego}' \sim p_{data}(I_{ego},I_{ego}')}[\mid \mid I_{ego} - I_{ego}' \mid \mid _1],
\end{equation}

\noindent where, $I_{ego}' = G(I_{exo})$.
The objective function for our network is the sum of conditional GAN loss in Eq. \eqref{eq_gan} and $L1$ loss in Eq. \eqref{eq_l1}, as represented in Eq. \eqref{eq_comb}: 
\vspace{-5pt}
\begin{equation}\label{eq_comb}
L_{network} = L_{cGAN} (G,D) + \lambda L_{L1}(G),
\end{equation}
where, $\lambda$ is the balancing factor between the losses.

\begin{figure}[t!]
\begin{center}
\includegraphics[width=\linewidth]{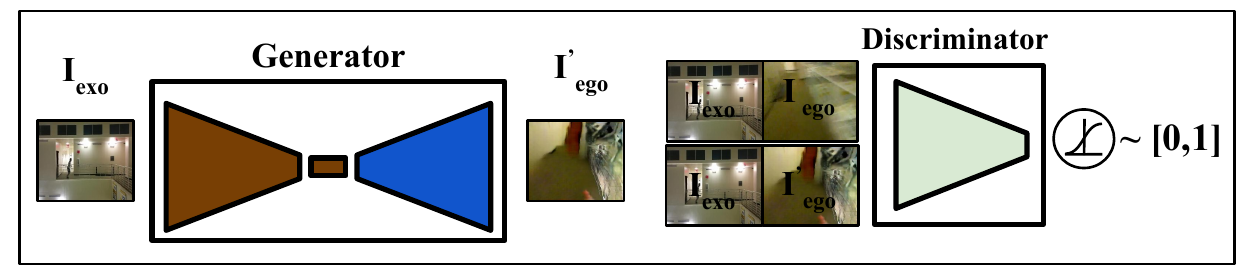}
\vspace{-20pt}
\caption{Image synthesis framework. An egocentric image is generated conditioned on an exocentric image. The exocentric image along with the real and synthesized egocentric images are passed to the discriminator as positive and negative pairs respectively.}
\label{fig:frameSynthesis}
\end{center}
\end{figure}

The architecture of our image generation network is shown in Fig. \ref{fig:frameSynthesis}. $I_{exo}$ is an exocentric image fed as a conditioning input to the network. The output of the generator is $I'_{ego}$ which is the generated image in egocentric domain. The discriminator is provided with the $(I_{exo},I'_{ego})$ pair as a negative example. The goal is to generate a $I'_{ego}$ realistic enough to be able to fool the discriminator. The real image pair, $(I_{exo},I_{ego})$ is also fed to the discriminator as a positive example. 

We utilize the baseline model of \cite{regmi2018cross} that was trained to generate street-view images from aerial images. We fine-tune the cross-view model for 15 epochs on our real and synthetic datasets. The images are first resized to 256 $\times$ 256 for generative tasks. We ran experiments with different hyperparameters but the ones from \cite{regmi2018cross} worked best. 

\subsection{Retrieval}
% \vspace{-5pt}
Given an egocentric video frame, we aim to retrieve its corresponding video frame across all the frames of all the exocentric videos. We perform retrieval based on the RGB values of the frames and also based on the optical flow. We perform retrieval using a two stream network with contrastive loss. We train a separate two stream network for RGB and one for Optical Flow. The architecture used for RGB based retrieval is shown in Fig. \ref{fig:retrievalNetworkArchitecture}. We use the same architecture for retrieval based on momentary optical flow (optical flow at that specific time frame), with an exception to the number of input channels (3 for RGB and 2 for optical flow). We extract view specific features from each stream and encourage a view invariant embedding by setting the difference between corresponding pairs to zero.
\vspace{-10pt}

\subsubsection{Optical Flow: }
We train a two stream network on the momentary optical flows extracted from each video. In others words, given a pair of simultaneously recorded exocentric and egocentric videos, we feed the optical flow at time t of the egocentric and exocentric video to the network as a positive pair. For any other pair of optical flow (frame $t_1$ in the egocentric and frame $t_2$ in exocentric where $t_1 \neq t_2$) the output of the network is set to 1 (negative pair). Since the optical flow maps are often very noisy, we perform a Gaussian smoothing over time in order to get more consistent flow maps, as a preprocessing step.

We train a network on the synthetic dataset (synthetic egocentric-exocentric pairs), and test it on the test set of the synthetic dataset. We perform the same experiment on the real dataset. We train and test another network on real dataset egocentric and exocentric pairs. We observe that the retrieval performance on the real data is not as favorable as the synthetic dataset, as the synthetic dataset is often less noisy, is in a more controlled environment, and has more training data. Given that the synthetic and real data are different in modality, we train a third retrieval network. We initialize the network with the weights trained on the synthetic dataset, and then fine-tune its convolutional layers on the synthetic data on the real data in order to benefit from the network pre-trained on the synthetic dataset. We observe that the retrieval performance of the fine-tuned network improves significantly on the real data.
\vspace{-10pt}
\subsubsection{RGB: }
We perform the same experiments on the raw RGB values of the two views. We use the same structure as before and follow the same fine-tuning paradigm to ensure a better learning using the synthetic-trained weights on the real data. Our experiments show a substantial retrieval quality for both of the real as well as synthetic data. As before, the network that is pre-trained on the synthetic dataset and fine-tuned on the real dataset yields the best retrieval performance on the real dataset.

\begin{figure}[t]
\begin{center}
\includegraphics[width=\linewidth]{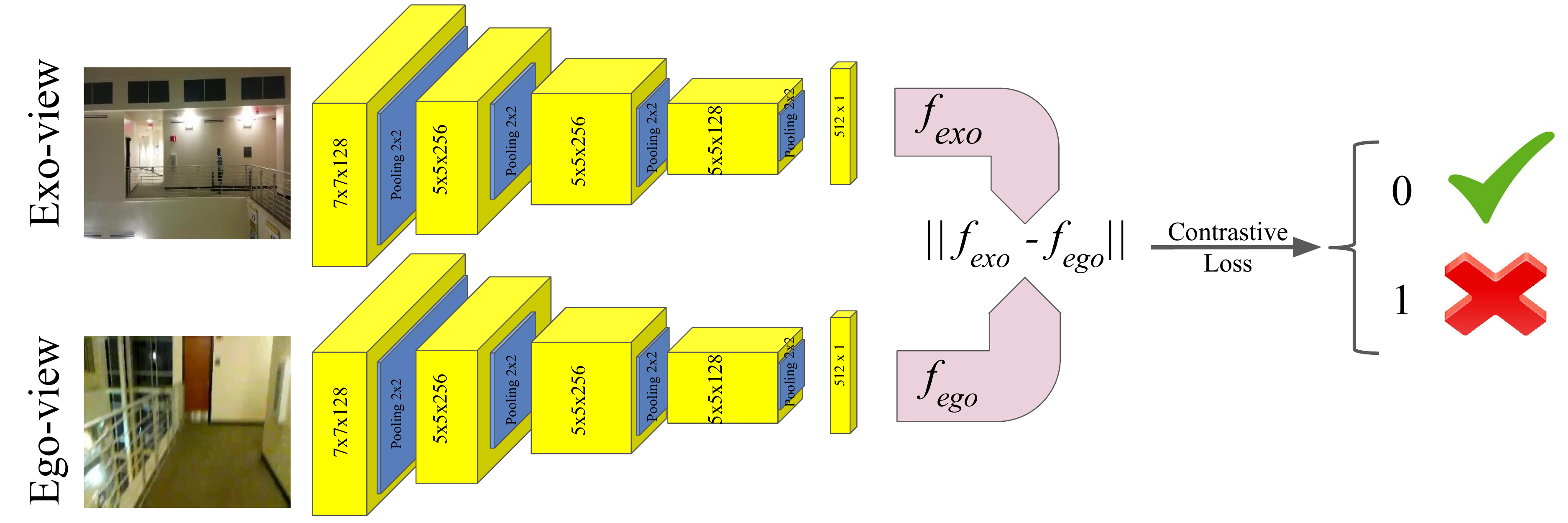}
\caption{Retrieval Network Architecture. }
\label{fig:retrievalNetworkArchitecture}
\end{center}
\end{figure}

\vspace{-10pt}
\section{Experiments}
\label{sec:experimentalResults}
%In the following, we evaluate our proposed approaches in terms of synthesis and retrieval. 

\subsection{Synthesis}
\begin{center}
\begin{figure*}[t!]
%\begin{subfigure}{0.95\textwidth}
\includegraphics[width=0.95\linewidth]{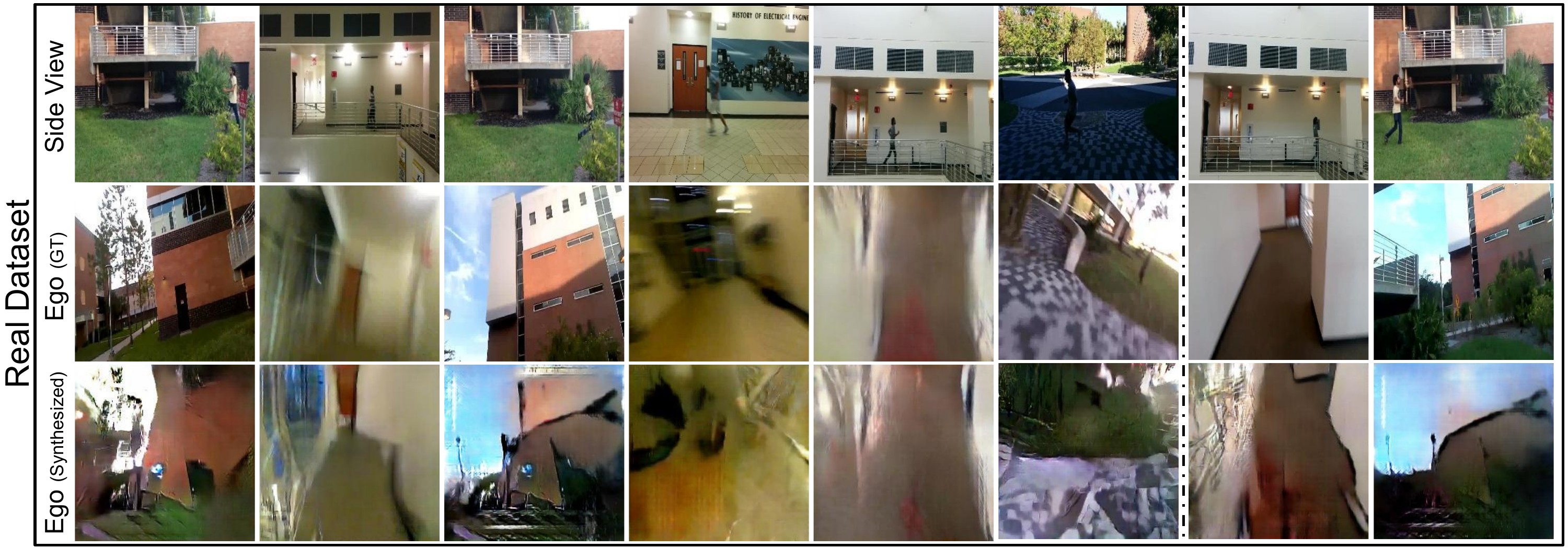}
\caption{Qualitative Results for synthesis on Real (upper block) and Synthetic Datasets (lower block). In each block, first row shows images in exocentric (side) view, second row shows their corresponding ground truth egocentric images and the third row shows egocentric images generated by our method.  }
\label{fig:synthesisQualitativeReal}
%\end{subfigure}
% \begin{subfigure}{1\textwidth}
\includegraphics[width=0.95\linewidth]{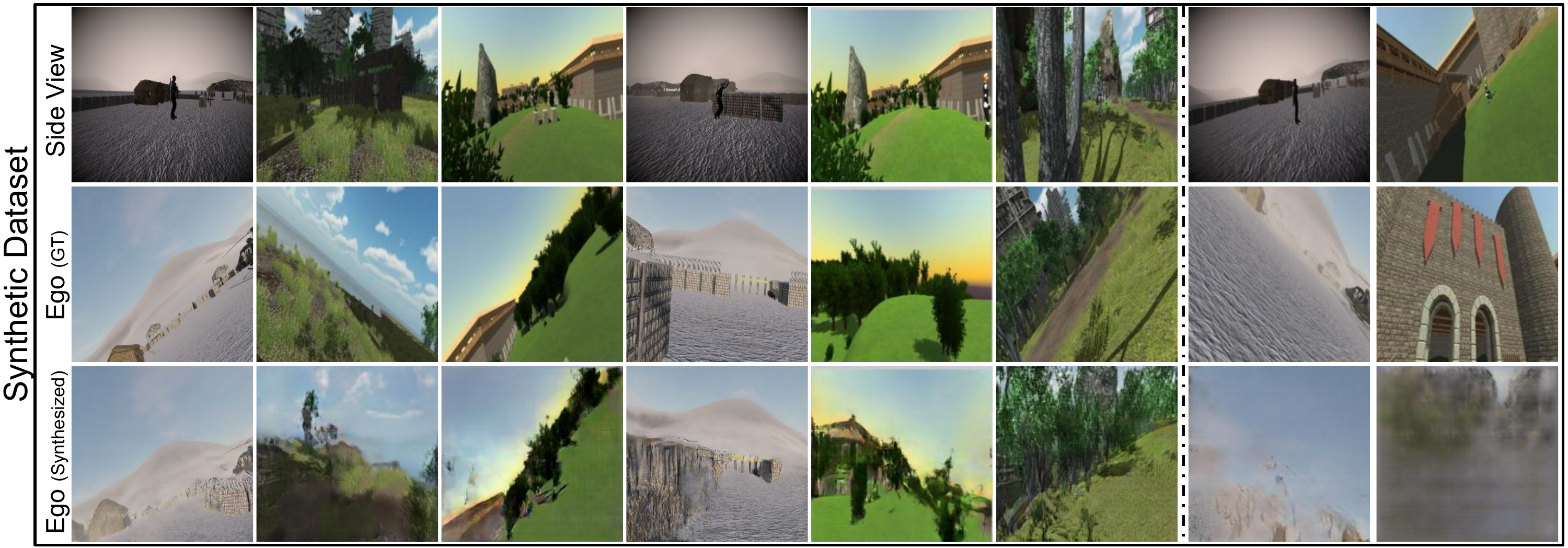}
% \caption{Qualitative Results for synthesis on Synthetic Dataset. }
% \label{fig:synthesisQualitativeSynthetic}
% \end{subfigure}
% \caption{Qualitative Results for synthesis on Real Dataset. }
% \label{fig:synthesisQualitativeReal}
\end{figure*}
\end{center}

A set of randomly selected qualitative results over real and synthetic datasets have been shown in Fig. \ref{fig:synthesisQualitativeReal}. The generated frames show that the network is successful at transforming the semantic information across the views. The generated images show blurriness for real dataset which is primarily because egocentric domain experiences motion in the frame rather than on the actor. The last two columns show some failure cases. The first failure case for real dataset shows the network is not able to learn the direction the person is facing so it is not able to generate the railings on right side of the person. The failure case for synthetic images show that the network is not able to hallucinate the textures in the scene. We use the following quantitative measures to evaluate the performance of the generated first person images:\\
\noindent\textbf{Inception Score~\cite{DBLP:conf/cvpr/SzegedyVISW16}:} measures the diversity of the generated samples within a class, and their representative of the class. The inception score is computed as the following: 
\begin{equation}
\label{eq:inceptionScore}
I = e^{E_xD_{KL}(p(y|x)||p(y))}
\end{equation}
where $x$ is a generated sample and $y$ is its predicted label.
We use the AlexNet model \cite{DBLP:journals/cacm/KrizhevskySH17} trained on Places dataset \cite{zhou2017places} with 365 categories to compute the inception score for images. 
% We also measure the inception score using the features extracted from the ego-stream of our RGB retrieval network trained on our synthetic and real datasets. 
Following the \cite{regmi2018cross}, we also compute inception scores on Top-1 and Top-5 classes, where Top-k means that top k predictions for each image are unchanged while the remaining predictions are smoothed by an epsilon equal to $\frac{1 - \Sigma top_k}{n-k}$.\\

\noindent\textbf{Structural-Similarity (SSIM):} measures the similarity between the images based on their luminance, contrast and structural aspects. SSIM values range between -1 and +1. A higher value means greater similarity between the compared images. It is computed as

\begin{equation}
SSIM(I_{ego}, I'_{ego})=\frac{(2\mu_{I_{ego}}\mu_{I'_{ego}}+c_1)(2\sigma_{I_{ego}I'_{ego}}+c_2)}{(\mu_{I_{ego}}^2 + \mu_{I'_{ego}}^2+c_1)(\sigma_{I_{ego}}^2 + \sigma_{I'_{ego}}^2+c_2)}
\end{equation}

\noindent\textbf{Peak Signal-to-Noise Ratio (PSNR):} measures the peak signal-to-noise ratio between two samples and evaluates the quality of the synthesized sample compared to the original sample. Higher values in PSNR imply better quality. It is computed as
\begin{equation}
PSNR(I_{ego},I'_{ego})=10log_{10}(\frac{max ^2I'_{ego}}{\frac{1}{n} \Sigma_{i=0}^{n}(I_{ego}[i]-I'_{ego}[i])^2})
\end{equation}
where $max I'_g = 255$ (maximum pixel intensity value).
\\
\noindent\textbf{Sharpness difference:} similar to \cite{journals/corr/MathieuCL15,regmi2018cross}, we compute the following:
\begin{equation}
SharpDiff(I_{ego},I'_{ego})=10log_{10}(\frac{max ^2I'_{ego}}{\frac{1}{N}\Sigma_{i}\Sigma_{j}|(\nabla_iY+\nabla_jY)-(\nabla_iY'+\nabla_jY')|})
\end{equation}
where the denominator corresponds to the difference between the gradients of the generated and ground truth image. Intuitively, we would like the difference between the gradients to be small.

\begin{table}[h]
 \small
  \centering
  \renewcommand{\arraystretch}{.9}
  \renewcommand{\tabcolsep}{.75mm} 
  \caption{\small Inception Scores for data and model distributions on Real and Synthetic Datasets.}
   \label{tab:inceptionScores}
   \begin{tabular*}{\textwidth}{l @{\extracolsep{\fill}} cccc}
        \toprule
        \multicolumn{1}{l}{\textbf{Images}} & \multicolumn{3}{c}{\textbf{Inception Score}}\\
       \cmidrule(lr){2-4}
       & all classes & Top-1 class & Top-5 classes\\
        \midrule
        Real Synthesized & $3.8280$ & $2.0315$& $3.4186$\\
    Real Ground-Truth &   $ 6.3787 $ & $2.6652$ & $5.2608$\\
 \midrule
    Synthetic Synthesized & $3.4320$ & $2.1045$& $3.5042$\\
    Synthetic Ground-Truth &   $ 4.5353 $ & $2.3815$ & $4.3695$\\
        \bottomrule
%         \vspace{-25pt}
    \end{tabular*}
\end{table}

\begin{table}[h]
 \small
  \centering
  \renewcommand{\tabcolsep}{1.5mm} 
  \caption{\small SSIM, PSNR and Sharpness Difference between real data and generated samples for Real and Synthetic Datasets.}
  \label{tab:ssim_psnr_sharpdiff}
    \begin{tabular}{lccc}
        \toprule
        \textbf{Dataset} & \textbf{SSIM} & \textbf{PSNR} & \textbf{Sharp Diff}\\
        \midrule
        Real & $0.4822$ & $18.1694$ &   $ 19.8142$\\
    Synthetic &   $ {0.5153} $ & $20.8976$ &  $ {20.5758} $
\\
        \bottomrule
%         \vspace{-25pt}
    \end{tabular}
\end{table}

The inception scores are shown in Table \ref{tab:inceptionScores}. The higher inception scores for the real dataset is expected as the network was pretrained on natural images (Places dataset). SSIM, PSNR and Sharpness Difference scores are reported in Table \ref{tab:ssim_psnr_sharpdiff}. All of the scores are higher for the Synthetic dataset compared to the real dataset. This is mainly due to the fact that the synthetic dataset has a controlled environment with less motion blur compared to egocentric frames in real dataset.

\subsection{Retrieval}
We evaluate the retrieval performance using the cumulative matching curve (CMC). The area under curve (AUC) of the curves are used as a quantitative measure. We evaluate retrieval using optical flow, and report the results in Fig. \ref{fig:retrievalCMC} (left). We also illustrate the retrieval results based on RGB in Fig. \ref{fig:retrievalCMC} (right). 

As explained before, we first train and test a two stream network on the synthetic dataset. The performance of this network is illustrated using the blue curve in Fig. \ref{fig:retrievalCMC}, and is referred to as train S test S, where S stands for synthetic data. The green curve shows the performance of the two stream retrieval network trained and tested on the real data (train R test R, where R stands for real dataset). The red and blue curves are not directly comparable as they are tested on different datasets (synthetic and real). In general, the retrieval performance is not high over the real dataset due to its less amount of data and high noise. The orange curve (train S test R), shows the retrieval performance of the network trained on the synthetic data directly on the real data, which generally does not perform better than the network trained on the real data (green). However, once we fine-tune the network trained on the synthetic data on the real data, we attain better performance (red curve, train S-R, test R). Except the blue curve which is tested on the synthetic data, all other curves are comparable as they have been tested on the real dataset. The best performance is achieved when the network is trained on synthetic data, and then its convolutional layers are tuned on the real data. The performance of chance (randomly ranking) is shown by the purple curve (chance).

\subsubsection{Retrieval based on Optical Flow:}
The cumulative matching curves for retrieval based on optical flow is shown in Fig. \ref{fig:retrievalCMC} (right). It can be observed that the network trained on synthetic and tested on real (orange) perform as chance level. The effect of adapting the synthetic network to the real data (red curve) is significant. As it can be observed the red curve (trained on synthetic, tuned on real data) does outperform the baselines on real data (green and orange curves). Please note that the blue curve has been evaluated on the synthetic data and therefore is not comparable to the other curves.

\subsubsection{Retrieval based on RGB:} 
\begin{figure}[t]
\begin{center}
\includegraphics[width=0.49\linewidth]{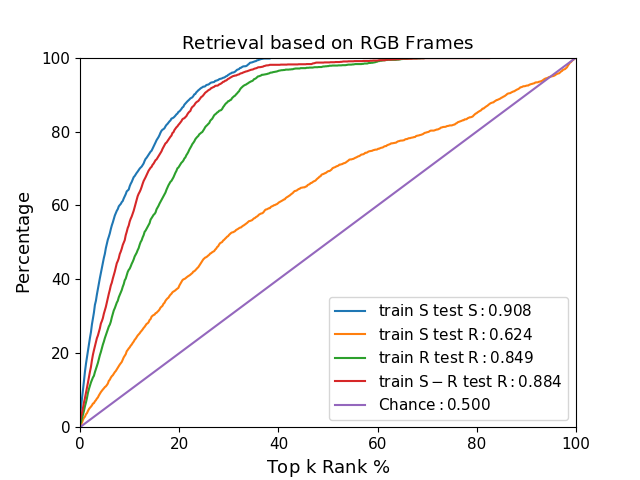}
\includegraphics[width=0.49\linewidth]{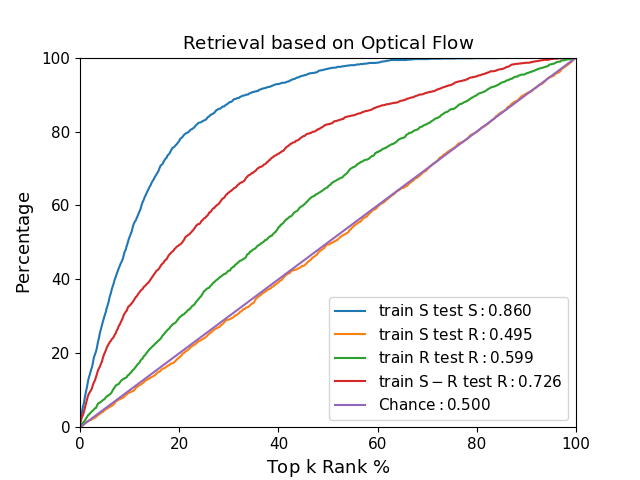}
\end{center}
\caption{Retrieval performance based on RGB (left) and optical flow (right). S stands for synthetic data and R stands for real data.}
\label{fig:retrievalCMC}
\end{figure}

The retrieval results based on RGB values are shown in Fig. \ref{fig:retrievalCMC} left. Similar to optical flow based retrieval, the phenomena of synthetic data being helpful in retrieving real data is observed. However, the improvement margin is less significant. This is due to the higher accuracy of the network trained on real data (green).   

% \begin{figure}
% \begin{center}
% \includegraphics[width=0.5\linewidth]{Rank_RGB.png}
% \end{center}
% \caption{Retrieval performance based on RGB frames. S stands for synthetic data and R stands for real data. It can be observed that generally retrieval on real data}
% \label{fig:retrievalRGB}
% \end{figure}

\begin{table*}[t]
 \small
  \centering
  \renewcommand{\tabcolsep}{0.5mm} 
  \caption{\small View Invariance-test based on Actions: In the synthetic dataset the chance level is $20\%$ as there are 5 action classes. In the real dataset the chance level is $12\%$ as there are 8 classes. }
  \label{tab:viewInvariance}
    \begin{tabular}{lccc}
        \toprule
        \textbf{Retrieval Network \textbackslash View} & \textbf{Egocentric View} & \textbf{Exocentric View} & \textbf{Both Views}\\
        \midrule
        train Synthetic OF & $37.71\%$ & $21.17\%$ &   $27.33\%$\\

        train Synthetic RGB & $29.05\%$ & $27.29\%$ &   $28.71\%$\\
        \midrule

    trained Real OF &   $33.49\%$ & $28.18\%$ &  $30.82\%$\\

trained Synthetic - Real OF &   $32.31\%$ & $32.97\%$ &  $30.72\%$\\

    trained Real RGB &   $42.58\%$ & $20.28\%$ &  $24.16\%$\\
trained Synthetic - Real RGB &   $ 42.58\% $ & $20.43\%$ &  $23.34\%$\\

\bottomrule
    \end{tabular}
\end{table*}

\begin{figure}[t]
\begin{center}
\includegraphics[width=0.49\linewidth]{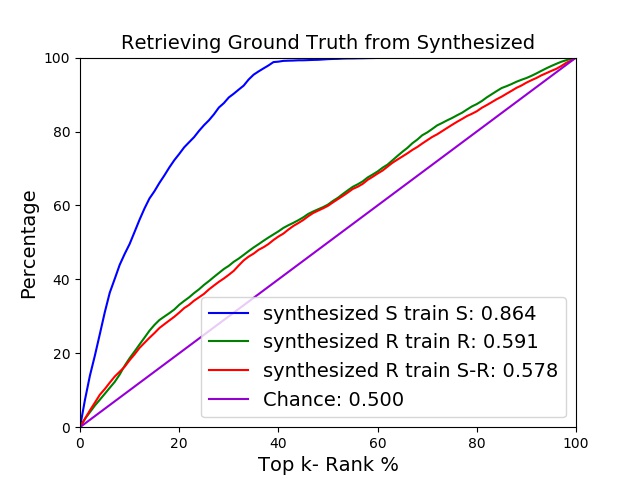}
\includegraphics[width=0.49\linewidth]{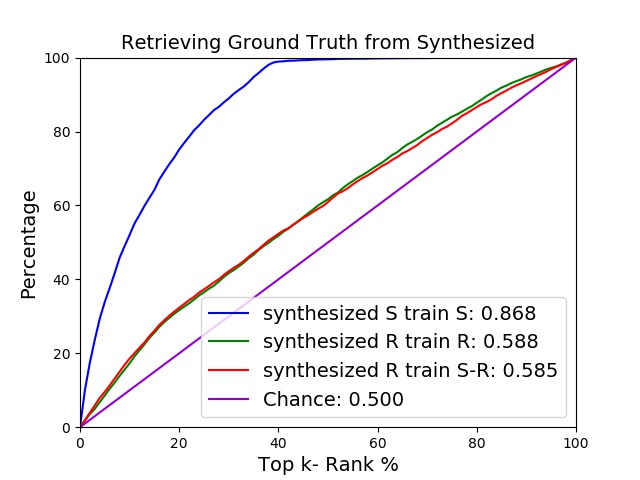}
\end{center}
\vspace{-15pt}
\caption{Retrieving the ground-truth egocentric, and exocentric images from the the synthesized images (left and right respectively). Similar to the figure \ref{fig:retrievalCMC}, S stands for synthetic data and R stands for real data. Synthetic synthesized and ground truth images are fed to the retrieval network trained on synthetic data (blue). The real (synthesized and ground-truth) egocentric images are fed to the networks trained on real data (green: trained on real and red: trained on synthetic and fine-tuned on real).}
\label{fig:cmc_synth_retrieval}
\end{figure}

\subsection{Retrieving Synthesized Images: }
As shown in Fig. \ref{fig:frameSynthesis}, given an exocentric image $I_{exo}$, the synthesis network outputs a synthesized image $I'_{ego}$, and the corresponding ground-truth egocentric frame is called $I_{ego}$. In this experiment, we explore if the synthesis preserves higher level information. In other words, is $I'_{ego}$ consistent with $I_{ego}$ and $I_{exo}$ in terms of high-level information? In order to answer this, we use the RGB retrieval network to extract egocentric features from the synthesized and ground truth egocentric images. In other words, we extract $f_{ego}(I'_{ego})$ and $f_{ego}(I_{ego})$ (where $f_{ego}$ and $f_{exo}$ are shown in Fig. \ref{fig:retrievalNetworkArchitecture}.). We store all the features extracted from all synthesized egocentric images in $F'_{ego}$, the features from the ground-truth egocentric images in $F_{ego}$, and the features extracted from the exocentric images in $F_{exo}$. 
For each synthesized egocentric image in $F'_{ego}$, we retrieve its corresponding ground truth exocentric feature from $F_{exo}$. The retrieval results are shown in Fig. \ref{fig:cmc_synth_retrieval} (left). We also retrieve its corresponding ground truth egocentric feature from $F_{ego}$. The results are shown in Fig. \ref{fig:cmc_synth_retrieval}. In both figures, the blue curve is the retrieval performance on the synthesized synthetic data, and the red and green curves show the retrieval on the synthesized real data using the different networks explained in the retrieval section.

\subsection{View-invariance Test}
Here we test the view-invariance of the retrieval network. To do so, we feed the training set (egocentric and exocentric RGB frames and optical flows) to our retrieval network and extract the features from their last fully connected layers (512 dimensions). In other words, we feed the egocentric frames to the ego stream and extract their features from the last fully connected layer. We train two separate SVM classifiers on the features extracted from each view of the retrieval network: one SVM on egocentric features and action labels, and another on exocentric actions and labels. We then evaluate the performance of each of the SVMs (reported in Table \ref{tab:viewInvariance} Egocentric view and exocentric view columns). A third SVM is then trained on all the features extracted from both views. In other words we pool all the features corresponding to each action independent of the fact that it is coming from the egocentric or exocentric stream. We then evaluate the performance of the third SVM on the first two. The classification performance of the SVM trained on both views does preserve the accuracy, and sometimes even outperforms the separately trained SVMs.

%\clearpage
\section{Discussion and Conclusion}
\label{sec:conclusion}
In this work we introduce new synthetic and real datasets of simultaneously recorded egocentric and exocentric videos. We show that performing tasks such as retrieval and synthesis from third person to first person are possible.
Future research can be done in the area of synthesis. Video generation is a possible extension of this effort. Also, embedding a view invariant representation in the bottleneck of the synthesis network can potentially unify the two tasks further. Other parameters such as camera and human pose and action labels can also be leveraged for better synthesis. Also, as we observed in our retrieval task, the synthetic data can be leveraged to address the lack of real data. We believe that this work provides useful datasets and baselines to address fundamental problems in relating first and third person images and videos.

{\small
\bibliographystyle{ieee}
\bibliography{sample}
}

\end{document}